\PassOptionsToPackage{unicode}{hyperref}
\PassOptionsToPackage{hyphens}{url}
\documentclass[
]{article}
\usepackage{amsmath,amssymb}
\usepackage{lmodern}
\usepackage{iftex}
\ifPDFTeX
  \usepackage[T1]{fontenc}
  \usepackage[utf8]{inputenc}
  \usepackage{textcomp} 
\else 
  \usepackage{unicode-math}
  \defaultfontfeatures{Scale=MatchLowercase}
  \defaultfontfeatures[\rmfamily]{Ligatures=TeX,Scale=1}
\fi
\IfFileExists{upquote.sty}{\usepackage{upquote}}{}
\IfFileExists{microtype.sty}{
  \usepackage[]{microtype}
  \UseMicrotypeSet[protrusion]{basicmath} 
}{}
\makeatletter
\@ifundefined{KOMAClassName}{
  \IfFileExists{parskip.sty}{%
    \usepackage{parskip}
  }{
    \setlength{\parindent}{0pt}
    \setlength{\parskip}{6pt plus 2pt minus 1pt}}
}{
  \KOMAoptions{parskip=half}}
\makeatother
\usepackage{xcolor}
\usepackage{longtable,booktabs,array}
\usepackage{calc} 
\usepackage{etoolbox}
\makeatletter
\patchcmd\longtable{\par}{\if@noskipsec\mbox{}\fi\par}{}{}
\makeatother
\IfFileExists{footnotehyper.sty}{\usepackage{footnotehyper}}{\usepackage{footnote}}
\makesavenoteenv{longtable}
\usepackage{graphicx}
\makeatletter
\def\maxwidth{\ifdim\Gin@nat@width>\linewidth\linewidth\else\Gin@nat@width\fi}
\def\maxheight{\ifdim\Gin@nat@height>\textheight\textheight\else\Gin@nat@height\fi}
\makeatother
\setkeys{Gin}{width=\maxwidth,height=\maxheight,keepaspectratio}
\makeatletter
\def\fps@figure{htbp}
\makeatother
\ifLuaTeX
  \usepackage{selnolig}  
\fi
\IfFileExists{bookmark.sty}{\usepackage{bookmark}}{\usepackage{hyperref}}
\IfFileExists{xurl.sty}{\usepackage{xurl}}{} 
\hypersetup{
  hidelinks,
  pdfcreator={LaTeX via pandoc}}

\author{}
\date{}

\begin{document}

\title{Deep Learning-Driven Inversion Framework for Shear Modulus Estimation in Magnetic Resonance Elastography (DIME)}

\author{
Hassan Iftikhar\textsuperscript{1,2}, 
Rizwan Ahmad\textsuperscript{1,3}, 
Arunark Kolipaka\textsuperscript{1,2,3}
}

\date{}  

\maketitle

\begin{center}
\textsuperscript{1}Biomedical Engineering, The Ohio State University, Columbus, Ohio, USA \\
\textsuperscript{2}Department of Radiology, The Ohio State University, Columbus, Ohio, USA \\
\textsuperscript{3}Davis Heart \& Lung Research Institute, The Ohio State University, Columbus, Ohio, USA \\
\vspace{0.5em}
\textbf{Email addresses:} \\
HI: \href{mailto:iftikhar.15@buckeyemail.osu.edu}{iftikhar.15@buckeyemail.osu.edu} \\
RA: \href{mailto:ahmad.46@osu.edu}{ahmad.46@osu.edu} \\
AK: \href{mailto:Arunark.Kolipaka@osumc.edu}{Arunark.Kolipaka@osumc.edu} \\
\vspace{0.5em}
\textbf{Corresponding Author:} Hassan Iftikhar
\end{center}
\newpage

\hypertarget{abstract}{%
\section{ABSTRACT}\label{abstract}}
The Multimodal Direct Inversion (MMDI) algorithm is widely used in
Magnetic Resonance Elastography (MRE) to estimate tissue shear
stiffness. However, MMDI relies on the Helmholtz equation, which assumes
wave propagation in a uniform, homogeneous, and infinite medium.
Furthermore, the use of the Laplacian operator makes MMDI highly
sensitive to noise, which compromises the accuracy and reliability of
stiffness estimates. In this study, we propose the Deep-Learning driven
Inversion Framework for Shear Modulus Estimation in MRE (DIME), aimed at
enhancing the robustness of inversion. DIME is trained on the
displacement fields-stiffness maps pair generated through Finite Element
Modelling (FEM) simulations. To capture local wave behavior and improve
robustness to global image variations, DIME is trained on small image
patches. We first validated DIME using homogeneous and heterogeneous
datasets simulated with FEM, where DIME produced stiffness maps with low
inter-pixel variability, accurate boundary delineation, and higher
correlation with ground truth (GT) compared to MMDI. Next, DIME was
evaluated in a realistic anatomy-informed simulated liver dataset with
known GT and compared directly to MMDI. DIME reproduced ground-truth
stiffness patterns with high fidelity (r = 0.99, R² = 0.98), while MMDI
showed greater underestimation. After validating DIME on synthetic data,
we tested the model in \emph{in vivo} liver MRE data from eight healthy
and seven fibrotic subjects. DIME preserved physiologically consistent
stiffness patterns and closely matched MMDI, which showed directional
bias. Overall, DIME showed higher correlation with ground truth and
visually similar stiffness patterns, whereas MMDI displayed a larger
bias that can potentially be attributed to directional filtering. These
preliminary results highlight the feasibility of DIME for clinical
applications in MRE.

\newpage
\hypertarget{introduction}{%
\section{1. INTRODUCTION}\label{introduction}}

Biomechanical tissue properties are closely linked to physiological and
pathological processes.\textsuperscript{1} Among these, stiffness has
been recognized as a valuable biomarker for detecting various
pathological states.\textsuperscript{2} Changes in tissue stiffness can
indicate the presence of diseases such as
fibrosis,\textsuperscript{3,4,5} tumors,\textsuperscript{6,7} or
inflammatory conditions.\textsuperscript{8} Historically, physicians and
clinicians have relied on palpation as a fundamental procedure to detect
the stiffness changes.\textsuperscript{9} However, palpation applies to
organs close to the body\textquotesingle s surface and cannot be applied
to the deeper tissues, making it challenging to detect abnormalities
non-invasively. Additionally, the diagnostic reliability of palpation is
limited due to its subjective nature.

Magnetic Resonance Elastography (MRE)\textsuperscript{10} has emerged as
a promising technique to estimate the stiffness of tissues
non-invasively. MRE is a three-step process in which (1) mechanical
waves are induced in the tissue of interest via an acoustic vibrator;
(2) the propagation of wave in the tissue of interest is captured using
phase-contrast magnetic resonance imaging\textsuperscript{11,12} with
motion-encoding gradients synchronized with the external vibration; and
(3) the wave images are processed to estimate the stiffness of the
tissue by solving an inverse problem.\textsuperscript{13}

Various inversion algorithms have been developed over time, including,
but not limited to, variations of Direct Inversion
(DI),\textsuperscript{14,15} Local Frequency Estimation
(LFE),\textsuperscript{16,17} Multimodal Direct Inversion
(MMDI),\textsuperscript{18} and Non-Linear Inversions
(NLI).\textsuperscript{19} These methods assume that the tissue of
interest is infinite, uniform, homogeneous, and isotropic. In practice,
however, these assumptions do not hold in most tissues, leading to
potentially inaccurate stiffness measurements and limiting the
robustness of stiffness estimation. Accurate estimation of tissue
stiffness is essential, as stiffness is not only a key biomarker for
staging liver fibrosis, but also a confounded marker influenced by other
histopathological conditions such as inflammation, portal hypertension,
and hepatic venous congestion.\textsuperscript{20} Precise stiffness
estimation, therefore, not only improves the reliability of fibrosis
staging but also enables disease-specific interpretation, allowing
physicians to better distinguish among multiple pathological processes
using stiffness as a single surrogate biomarker.

Recently, deep learning (DL)-based methods have gained prominence for a
variety of imaging tasks, including classification, segmentation,
registration, and reconstruction.\textsuperscript{21,22,23,24} To
address the above-discussed limitations in MRE inversion, Murphy et
al.\textsuperscript{25} proposed a DL-based framework to estimate tissue
stiffness from displacement wavefields; however, this work relied on
simulated wavefields that were created by assuming homogeneous material
properties throughout the geometry of the tissue, neglecting the local
gradients in mechanical properties. Subsequently, Solamen et
al.,\textsuperscript{26} demonstrated MRE displacement-to-elastogram
mapping using Convolutional Neural Networks (CNNs), but the performance
from their approach was limited by insufficient training diversity and
lack of noise considerations. Scott et al.,\textsuperscript{27}
attempted to address the homogeneity assumptions by introducing
inhomogeneous material distributions using a coupled harmonic oscillator
model. However, this dataset did not fully capture the elastic waves
physics or boundary conditions essential for accurate stiffness
estimation. Also, the use of curl as a preprocessing step is known to
amplify the noise. Ma et al.\textsuperscript{28} introduced a
dual-consistency framework and utilized Travelling Wave Expansion (TWE)
model for dataset generation, but this required multifrequency and
multidirectional MRE acquisitions, hence increasing the scan time while
still relying on simplifying assumptions of tissue uniformity, linearity
and homogeneity. Ragoza et al.\textsuperscript{29} proposed the utility
of physics-informed neural networks to solve this inverse problem;
however, the inversion still relied on Helmholtz inversion equation,
which assumes tissue linearity and homogeneity. Zhang et
al.\textsuperscript{30} utilized finite difference time domain method
for synthetic data generation; however, their framework did not
explicitly account for realistic boundary conditions. Recently, Bustin
et al.\textsuperscript{31} proposed ElastoNet, a CNN--based inversion
framework trained on synthetically generated 5×5 wave patches derived
from point-source excitations, achieving frequency and resolution
independent stiffness mapping under locally homogeneous, isotropic, and
viscoelastic material assumptions. To overcome the inherent limitations
of analytical MRE inversion, most notably their reliance on assumptions
of tissue uniformity, homogeneity, isotropy, and an infinite medium,
there remains a need for an inversion framework that can more accurately
model realistic tissue behavior. DL offers a promising data-driven
alternative. However, DL-based methods often require a large corpus of
labeled training data. In many applications, including MRE, such
ground-truth data is not available. Moreover, the performance of
DL-based models is fundamentally constrained by the realism and quality
of the training data, making accurately modeled datasets essential for
reliable stiffness estimation. Although self-supervised and unsupervised
learning strategies have been proposed to mitigate this dependency, they
typically require de novo training for each dataset, which limits their
clinical utility.\textsuperscript{32} In this work, we propose a
\textbf{D}eep Learning-based \textbf{I}nversion framework for shear
\textbf{M}odulus Estimation in MR\textbf{E} \textbf{(DIME)} that is
designed to overcome these challenges while preserving clinical
practicality. DIME introduces a pipeline for generating realistic paired
displacement--stiffness data through FEM, capable of capturing spatially
varying tissue mechanics that conventional simplified models fail to
represent. The dataset is constructed in a patch-based format, motivated
by the fact that stiffness is a local mechanical property, and the
corresponding shear wavelength varies spatially with local stiffness. A
CNN is then used to learn the mapping from complex-valued displacement
field patches to their corresponding shear modulus. By training on
patches rather than full images, DIME effectively learns localized
wavelength--stiffness relationships, enabling accurate stiffness
estimation at fine spatial scales, which is a key advantage of the
framework. To evaluate the performance of DIME, we developed a highly
realistic and innovative dataset, generated through advanced simulation
techniques.\textsuperscript{33} This synthetic dataset not only captures
the complexities of tissue behavior but also provides a robust
foundation for assessing the performance of DIME against existing
methods like MMDI. By using this carefully designed dataset, DIME is
evaluated under conditions that closely approximate real-world tissue
mechanics, where ground-truth stiffness is typically unknown, allowing
direct performance assessment of both DIME and MMDI against known
reference values. Once trained, DIME performs patch-wise stiffness
estimation within milliseconds, making it computationally efficient and
practically deployable.

\hypertarget{theory}{%
\section{2. THEORY}\label{theory}}

DIME formulates the problem of stiffness estimation as a supervised
learning task, offering a data-driven alternative to analytical
inversion approaches. Rather than relying on derivative-based
formulations such as the Helmholtz equation, DIME directly learns the
mapping between complex displacement wavefields and spatially varying
stiffness distributions using a CNN \(f_{\theta}\), parameterized by
weights $\theta:
$
\[
f_{\theta} : \mathbf{u} \;\rightarrow\; \mu
\tag{1}
\]

Where \(\mathbf{u} \in \mathbb{R}^{2xHxW}\) contains two real-valued
channels corresponding to the real and imaginary components of the
complex displacement field, and \(\mu \in \mathbb{R}^{HxW}\) is the
shear modulus map. The high-level description of DIME is provided in Fig
1.

\begin{figure}[htbp]
    \centering
    \includegraphics[width=\textwidth]{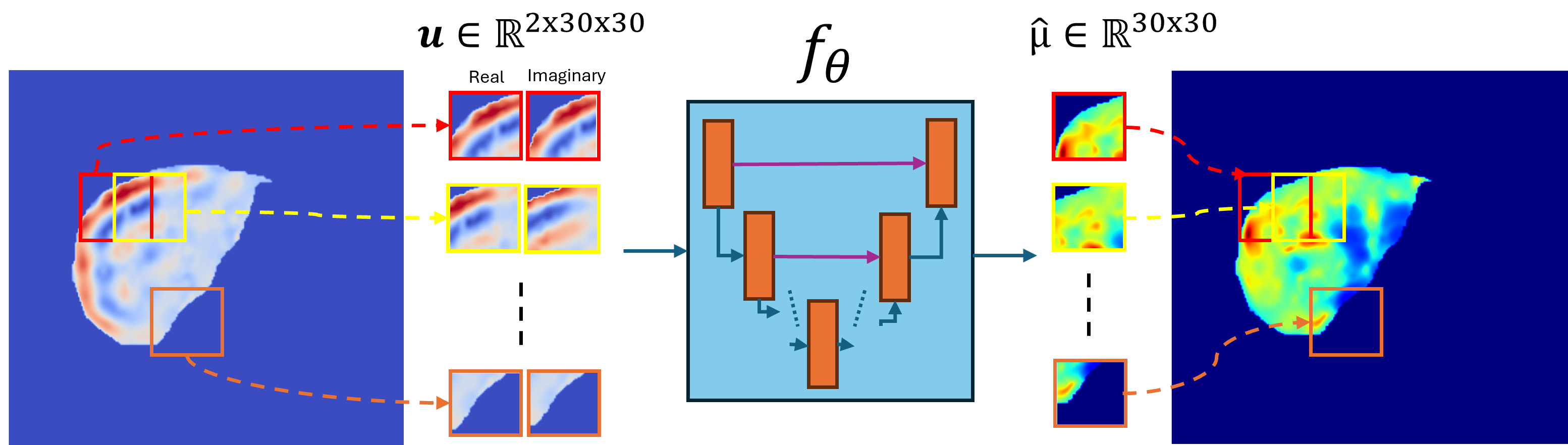}
    \caption{Overview of DIME Framework: Complex displacement
fields are divided into 2×30×30 patches (real and imaginary channels)
and processed by a neural network\hspace{0pt} \(f_{\theta}\) to predict
corresponding stiffness patches \(\mu \in \mathbb{R}^{30x30}\), which
are then aggregated downstream in the pipeline into a full stiffness
map.}
    \label{fig:your-label}
\end{figure}

\hypertarget{neural-network-architecture}{%
\subsection{\texorpdfstring{2.1. Neural Network Architecture
}{2.1. Neural Network Architecture }}\label{neural-network-architecture}}

The proposed model (\(f_{\theta}\)) is based on a U-Net, designed to map
complex-valued displacement fields to shear modulus distributions. Each
input is a 2×30×30 patch containing the real and imaginary components of
the first harmonic displacement field, and the network outputs a
corresponding 30×30 stiffness map. The network consists of four encoding
and four decoding blocks, with symmetric skip connections between
corresponding levels. Each convolutional block uses 3×3 kernels and
Leaky ReLU activations, followed by batch normalization. The number of
channels starts at 128 in the first layer and doubles at each subsequent
block (i.e., 64 → 128 → 256 → 512), before symmetrically decreasing back
to 1 in the final output layer. The architecture of the model is
described in Fig 2.

\begin{figure}[h]
    \centering
    \includegraphics[width=\linewidth]{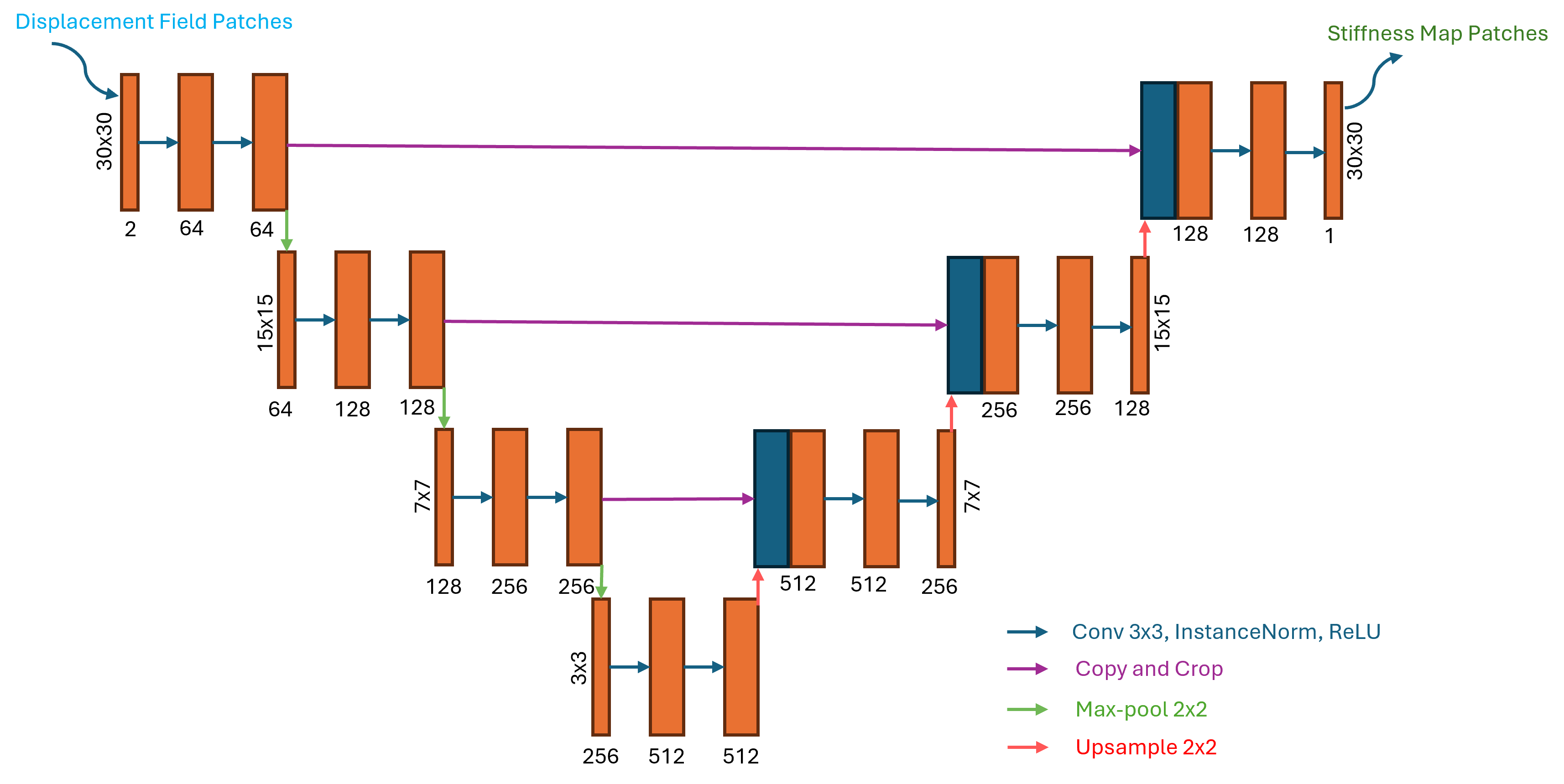}
    \caption{Model architecture}
    \label{fig:your-label}
\end{figure}

\hypertarget{dataset-generation}{%
\subsection{2.2. Dataset Generation}\label{dataset-generation}}

To train \(f_{\theta}\), we constructed a dataset of 900 synthetic
phantoms generated via Finite Element Modeling (FEM) using COMSOL
Multiphysics (COMSOL Inc., Massachusetts, USA). These phantoms were
designed to encompass a wide range of anatomically and mechanically
realistic patterns through variations in geometry, material properties,
inclusion configurations, and boundary conditions.

All simulations were performed with a Poisson's ratio of 0.499 and a
density of 1000~kg/m³. Planar wave propagation was induced by applying
prescribed displacements ranging from 0.3~mm to 0.9~mm. Representative
examples from key phantom classes are illustrated in Fig. 3\textbf{.}
All material regions (including inclusions and background tissue) were
assigned shear modulus values sampled from a uniform distribution of 1
to 8\,kPa and damping from 0.05 to 0.3. Circular phantoms were modeled
in 3D and square phantoms in 2D, with mesh resolution adapted
accordingly. The complete dataset composition is summarized in Table 1.

\begin{figure}[htbp]
    \centering
    \includegraphics[width=\textwidth]{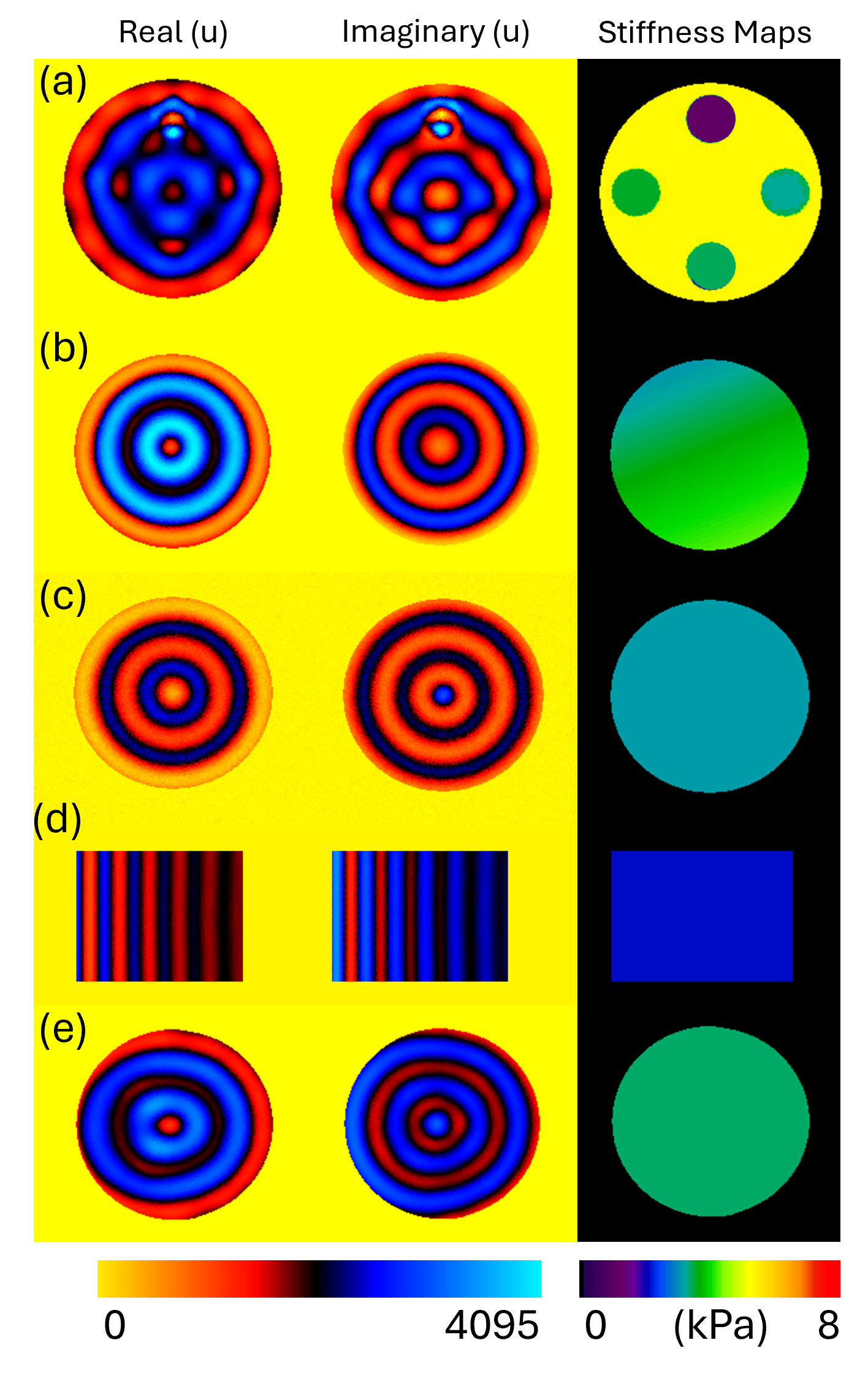}
    \caption{Simulated phantoms used for stiffness reconstruction.
(a) Phantom with multiple circular inclusions of varying stiffness. (b)
Phantom with linear stiffness variations. (c) Homogeneous circular
phantom with centered driver. (d) Homogeneous square phantom (e)
Homogeneous circular phantom with off-centered driver. Each set shows
the real and imaginary components of displacement fields along with the
corresponding stiffness map.}
    \label{fig:phantoms}
\end{figure}

\begin{table}[htbp]
\centering
\caption{Specifications of FEM-based simulations}
\label{tab:fem_specs}

\begin{tabular}{p{0.30\textwidth} p{0.10\textwidth} p{0.08\textwidth} p{0.34\textwidth} p{0.13\textwidth}}
\toprule
\textbf{Phantom Type} & \textbf{Dim.} & \textbf{Count} & \textbf{Description} & \textbf{DOF} \\
\midrule
Homogeneous cylindrical phantoms (Fig.~3c) & 3D & 400 &
Uniform stiffness, centered planar wave excitation &
248{,}205 \\

Cylindrical phantom with linear stiffness variations (Fig.~3b) & 3D & 100 &
Linearly varying $\mu$, with centered excitation &
248{,}205 \\

Cylindrical phantoms with off-centered excitation (Fig.~3e) & 3D & 100 &
Uniform $\mu$, with asymmetric boundary displacement &
457{,}269 \\

Cylindrical phantoms with 4 fixed-position inclusions (Fig.~3a) & 3D & 50 &
Four inclusions with fixed locations; varying stiffness and damping &
225{,}588 \\

Cylindrical phantoms with 2 random inclusions placement & 3D & 50 &
Two inclusions placed randomly per phantom &
252{,}801 \\

Homogeneous square phantoms (Fig.~3c) & 2D & 100 &
Uniform stiffness, planar wave propagation &
25{,}530 \\

Square phantoms with 4 random inclusions & 2D & 100 &
Four inclusions per phantom, randomly positioned and parameterized &
34{,}018 \\
\bottomrule
\end{tabular}
\end{table}

The complex-valued first harmonic displacement field was extracted as
the first harmonic of the temporal Fourier transform of simulated phase
images. Displacement and stiffness maps were then resampled to a uniform
spatial resolution of 1\,mm and divided into 30×30 patches.

Each input patch contained two real-valued channels corresponding to the
real and imaginary parts of the displacement field, while the output
patch represented the shear modulus. Patches containing fewer than 50\%
non-zero pixels were excluded, resulting in 25,000 training, 5,000
validation, and 3,000 testing patches. Validation and test sets were
drawn from the same distribution as the training data, without spatial
overlapping. This patch-based approach improved sampling efficiency and
enabled localized learning of wave-to-stiffness mappings across diverse
anatomical and mechanical configurations. Fig 1 illustrates the
patch-based approach.

\hypertarget{model-training}{%
\subsection{\texorpdfstring{2.3. Model Training
}{2.3. Model Training }}\label{model-training}}

The model was trained using the Adam optimizer (learning rate $3\times10^{-4}$,
batch size 128) for 150 epochs on an NVIDIA RTX 3090 GPU. A step-based
learning rate scheduler was used, with a decay factor ($\lambda = 0.8$) applied
every 20 epochs. The total training time was approximately 10 hours.
Early stopping based on validation loss was used to prevent overfitting.
The learning objective is to minimize the discrepancy between predicted
and ground-truth stiffness maps over a dataset of FEM-generated samples,
formulated as the following optimization problem:

\[
\theta^{*} = \arg\min_{\theta} \sum_{i=1}^{N} 
\mathcal{L}\big(f_{\theta}(u_i),\, \mu_i\big)
\tag{2}
\]

The composite loss function \(\mathcal{L}\) includes two components:

\[\mathcal{L = \ }\mathcal{L}_{MSE} + \lambda.\mathcal{L}_{Total\ Variation}\]

The term \(\mathcal{L}_{MSE}\) computes pixel-wise reconstruction error,
and is defined as

\[
L_{\text{MSE}}
= \frac{1}{N C H W}
\sum_{i=1}^{N} \sum_{c=1}^{C} \sum_{p=1}^{H} \sum_{q=1}^{W}
\left( \hat{\mu}_{i}^{(c)}(p,q) - \mu_{i}^{(c)}(p,q) \right)^{2}
\]

Where N, C, H, and W denote the batch size, number of channels, image
height, and width respectively. \({\widehat{\mu}}_{i}^{(c)}(p,q)\) and
\(\mu_{i}^{(c)}(p,q)\) represent the ground truth and predicted shear
modulus at spatial location (p,q) for channel c of i\textsuperscript{th}
sample. The isotropic total variation term given by
\(\mathcal{L}_{Total\ Variation}^{iso}\) promotes spatial smoothness
within each patch, discouraging spurious noise and artifacts without
over-smoothing anatomical features. It is defined as:

\[\mathcal{L}_{Total\ Variation}^{iso} = \ \sum_{i = 1}^{N}{\sum_{c = 1}^{C}{\sum_{p = 1}^{H}{\sum_{q = 1}^{W}{\ \ \sqrt{{{(\nabla}_{x}{\widehat{\mu}}_{i}^{(c)}(p,q))}^{2} + {{(\nabla}_{y}{\widehat{\mu}}_{i}^{(c)}(p,q))}^{2} + \epsilon}}}}}\]

Where \(\nabla_{x}\ \)and \(\nabla_{y}\) denote the forward finite
differences along the horizontal and vertical directions, respectively,
and \(\epsilon = 10^{- 8}\) is a small constant added for numerical
stability. The hyperparameter $\lambda$ controls the strength of this
regularization. The regularization coefficient $\lambda$ balances fidelity and
smoothness, helping patch-wise smoothness while preserving structural
features relevant to anatomical variations. The code of DIME will be
publicly available with some evaluation datasets upon publication at
https://github.com/HassanIftikhar013/DIME.

\hypertarget{methods}{%
\section{3. METHODS}\label{methods}}

To evaluate the generalization capability of DIME, we designed a series
of experiments using datasets distinct from the training set and
arranged them in the order of increasing complexity. The goal was to
assess how well a model trained on small synthetic patches generalizes
to larger, full-sized phantoms and in vivo images using a patch-by-patch
inference approach (Fig. 1). We first tested the model in homogeneous
and heterogeneous digital phantoms to evaluate its accuracy in detecting
varying stiffness levels. Next, we used a hybrid dataset by embedding
anatomical heterogeneity from in vivo MMDI scans into new FEM
simulations. Finally, we applied the trained model to in vivo liver MRE
data to assess real-world performance. In all evaluations, we
benchmarked DIME's performance against MMDI, the current gold standard
in stiffness reconstruction, to assess how well our data-driven approach
compares to traditional inversion techniques.

\hypertarget{evaluation-of-dime-on-simulated-phantoms}{%
\subsection{3.1. Evaluation of DIME on simulated
phantoms}\label{evaluation-of-dime-on-simulated-phantoms}}

\emph{(a) Homogeneous Phantom Evaluation:} To evaluate DIME on
previously unseen data, we generated a new set of 50 homogeneous
cylindrical phantoms specifically for testing. This dataset was distinct
from the training set and was designed to assess the model's ability to
recover uniform stiffness values from simple geometries. The phantoms
were simulated using the same FEM framework described in Section 2.2,
with constant shear moduli sampled from 1 to 8 kPa. All other simulation
parameters (e.g., damping, density, vibration frequency) matched those
used during training. First harmonic displacement fields were extracted
and divided into overlapping 20×20 patches (stride = 3). Patch-wise
predictions were aggregated by averaging overlaps to reconstruct
full-field stiffness maps.

For comparison, MMDI maps were computed on the same data.
Post-processing involved Butterworth bandpass filtering
(2--128\,waves/FOV) to remove longitudinal waves and directionally
filtered to remove reflected waves. This evaluation provided a
controlled baseline to benchmark DIME's accuracy against MMDI under
minimal spatial variation. Evaluation metrics included mean stiffness,
inter-pixel standard deviation, and R\textsuperscript{2} relative to the
known ground truth.

\emph{(b) Heterogeneous Phantom Evaluation:} To test the sensitivity of
DIME to various stiffness regions in a geometry, we evaluated DIME on
cylindrical phantoms containing four randomly placed inclusions. These
phantoms were constructed using the same simulation framework as before,
but with spatially varying material properties sampled from 1 to 8 kPa
for both background and inclusions. Inclusion shapes, locations, and
stiffness values were randomized across the dataset. Notably, this
specific pattern of heterogeneity was not present in the training set,
making it a strong test of generalization. As before, testing was
performed in a patch-wise manner, and full stiffness maps were
reconstructed by aggregating patch predictions.

Predicted maps were evaluated for their ability to localize inclusions,
preserve structural boundaries, and recover regional stiffness values,
with results benchmarked against MMDI. MMDI-based stiffness maps were
calculated by applying the same post-processing as described in Section
3.1 (a).

\hypertarget{evaluation-of-dime-on-anatomy-informed-phantoms}{%
\subsection{3.2. Evaluation of DIME on anatomy-informed
phantoms}\label{evaluation-of-dime-on-anatomy-informed-phantoms}}

To assess the performance of DIME on realistic and spatially
heterogeneous cases, we developed a novel hybrid dataset that mimics the
complexity of in vivo liver tissue while retaining access to
ground-truth stiffness. This was achieved by first acquiring liver MRE
data from 15 human subjects using a 1.5T clinical scanner (Aera, Siemens
Healthineers, Erlangen, Germany), after obtaining IRB approval and
written informed consent. Each subject was positioned in a head-first
supine position inside the scanner, and an acoustic driver (Resoundant
Inc., Rochester, MN) was used to generate the 60 Hz mechanical waves in
the liver. A gradient echo-based MRE sequence\textsuperscript{34,35} was
used to obtain shear waves in the axial slices of the liver. Imaging
parameters included TR/TE of 25/20.8\,ms, field of view of
36\,×\,36\,cm², matrix size of 128\,×\,64 with GRAPPA acceleration of 2,
and four contiguous 5\,mm axial slices with motion encoding applied in
the slice direction (duration: 16.7\,ms). Four temporal phase offsets
were acquired per slice to capture the propagating wavefield. Wave
images were processed using the MMDI algorithm, applying the same
post-processing pipeline described in Section 3.1, including Butterworth
bandpass filtering (2--128\,waves/FOV) and directional filter in 4
directions. The resulting stiffness maps were then smoothened using the
polynomial fitting method proposed by Iftikhar et
al.\textsuperscript{33} to generate continuous, spatially heterogeneous
distributions.

These smooth stiffness maps were used as inputs for FEM simulations,
which generated corresponding displacement fields. The resulting hybrid
dataset allowed for a realistic evaluation of DIME's performance on
wavefields closely resembling in vivo data, while maintaining a known
ground-truth stiffness distribution for quantitative comparison.

The final dataset consisted of 50 slices extracted from 15 subjects,
with the FEM-generated displacement fields serving as the model input,
and the polynomial-fitted MMDI stiffness maps serving as the ground
truth outputs.

\hypertarget{evaluation-of-dime-on-in-vivo-settings}{%
\subsection{3.3. Evaluation of DIME on in vivo
settings}\label{evaluation-of-dime-on-in-vivo-settings}}

For in vivo evaluation, we applied the trained DIME model directly to
the acquired wave images from the 15 subjects described in Section 3.2.
Unlike previous evaluations that relied on FEM-generated displacement
data, this experiment assessed the model's performance on wavefields
acquired in vivo subjects. The displacement data was processed
patch-by-patch using DIME to generate full-field stiffness maps. These
predictions were then compared to the corresponding MMDI-derived
stiffness maps, which served as the reference standard. To ensure
consistency, MMDI maps were processed with the same post-processing
pipeline described earlier in Section 3.2.

\hypertarget{results}{%
\section{4. RESULTS}\label{results}}

\hypertarget{validation-of-dime-on-simulated-phantoms}{%
\subsection{4.1. Validation of DIME on simulated
phantoms}\label{validation-of-dime-on-simulated-phantoms}}

\emph{(a) Validation on Homogeneous Phantoms:} Fig. 4 summarizes the
reconstruction performance of DIME and MMDI across 50 simulated
homogeneous phantoms. The real and imaginary components of the input
wavefield are shown along with the predicted stiffness maps from both
methods. Ground-truth (GT), DIME-derived, and MMDI-derived mean
stiffness values (in kPa) are indicated. Quantitatively, DIME
consistently produced stiffness maps with low inter-pixel variability
and mean values closely aligned with ground truth. In contrast, MMDI
reconstructions exhibited higher spatial variation and overestimated
stiffness in most cases. The mean ± standard deviation across phantoms
(Fig. 4b) shows that DIME maintained tight agreement with the true
values, while MMDI showed larger deviations. Fig. 4c shows the
correlation plot between DIME VS GT and MMDI vs GT. DIME achieved a high
correlation with ground truth, while MMDI displayed a lower correlation
and significant bias across the range of stiffness values.

\begin{figure}[htbp]
    \centering
    \includegraphics[width=\textwidth]{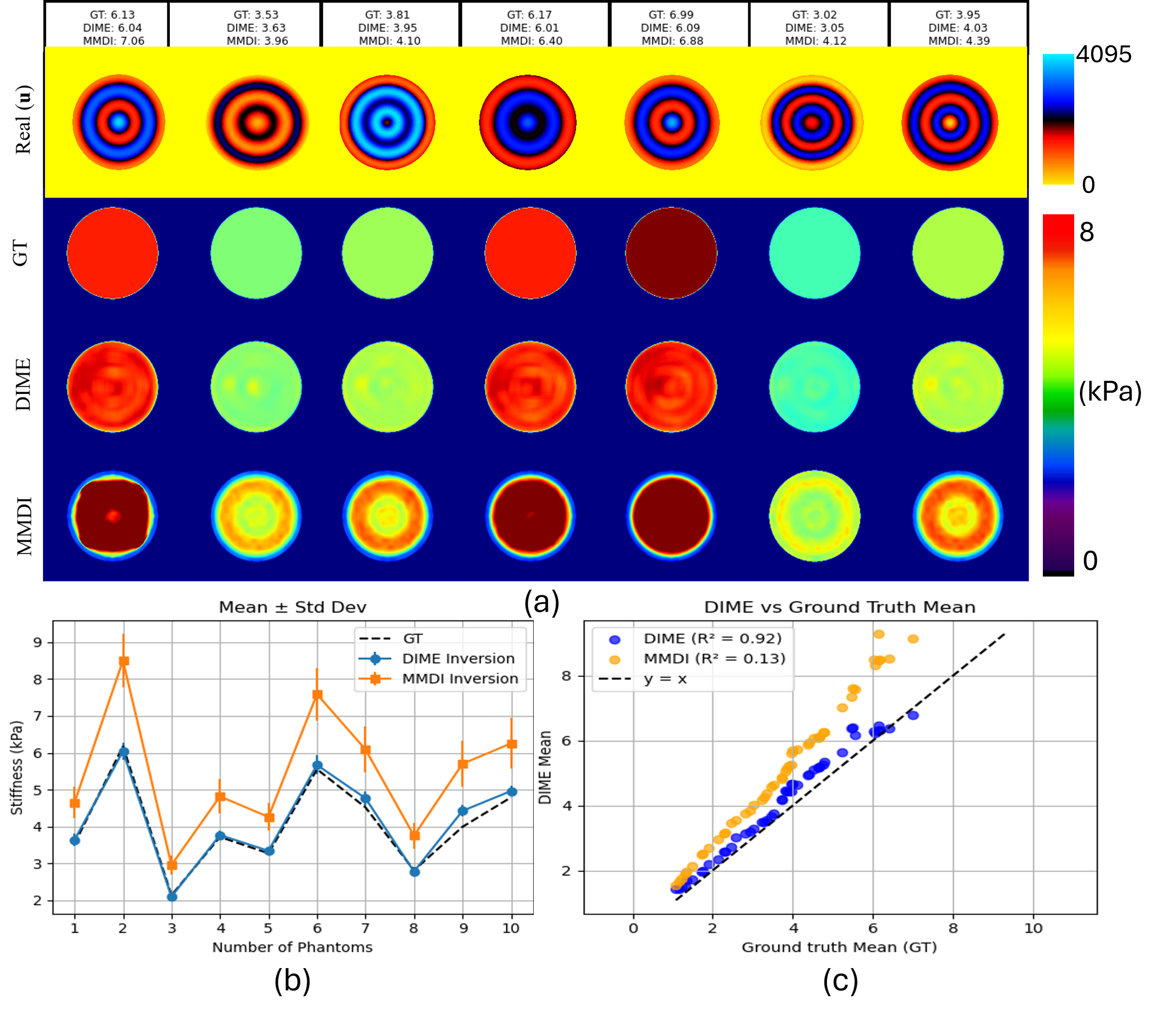}
    \caption{
            (a) Representative examples for Real wavefield components and
            corresponding stiffness maps reconstructed using DIME and MMDI. Mean
            stiffness values (in kPa) are shown above each column. \textbf{(b)} Mean
            ± standard deviation of predicted stiffness for DIME and MMDI compared
            to ground truth (GT) for 10 representative phantoms \textbf{(c)}
            Predicted vs. ground-truth mean stiffness for entire dataset; MMDI shows
            high bias, whereas DIME demonstrates reasonable alignment with ground
            truth.
    }
    \label{fig:yourlabel}
\end{figure}

\emph{(b) Validation on Heterogeneous Phantoms:} Fig. 5 demonstrates a
comparison of DIME and MMDI reconstructions on representative
heterogeneous phantoms containing multiple inclusions of varying
stiffness. Visual inspection indicated that DIME more accurately
delineates inclusion boundaries and preserves structural contrast
compared to MMDI, which often smooths transitions and introduces shape
distortion. Across examples, DIME consistently localized inclusions and
maintained sharper transitions between regions of varying stiffness.

Quantitative evaluation is summarized in Fig. 6a, which reports
region-wise mean and standard deviation comparisons across methods. DIME
showed strong agreement with ground-truth. In contrast, MMDI exhibited
systematic overestimation, especially in high-stiffness inclusions, with
greater intra-region variance. DIME showed more stable predictions
across background and inclusion regions, indicating better
generalization to complex spatial distributions. Bland--Altman analysis
(Fig. 6b) further supports the comparative evaluation of DIME and MMDI.
When compared against ground truth, DIME exhibited narrower limits of
agreement (-0.27 to 0.76\,kPa) and a smaller bias of 0.25\,kPa. In
contrast, MMDI showed wider limits of agreement (0.05 to 1.78\,kPa) and
a larger bias of 0.91\,kPa when estimating the overall mean stiffness of
the phantoms. Similar trends were observed in region-wise analysis,
where individual ROIs reinforced that MMDI consistently overestimated
stiffness, particularly in regions with higher ground-truth stiffness
values, indicating a systematic bias that was not seen in DIME.

\begin{figure}[htbp]
    \centering
    \includegraphics[width=\textwidth]{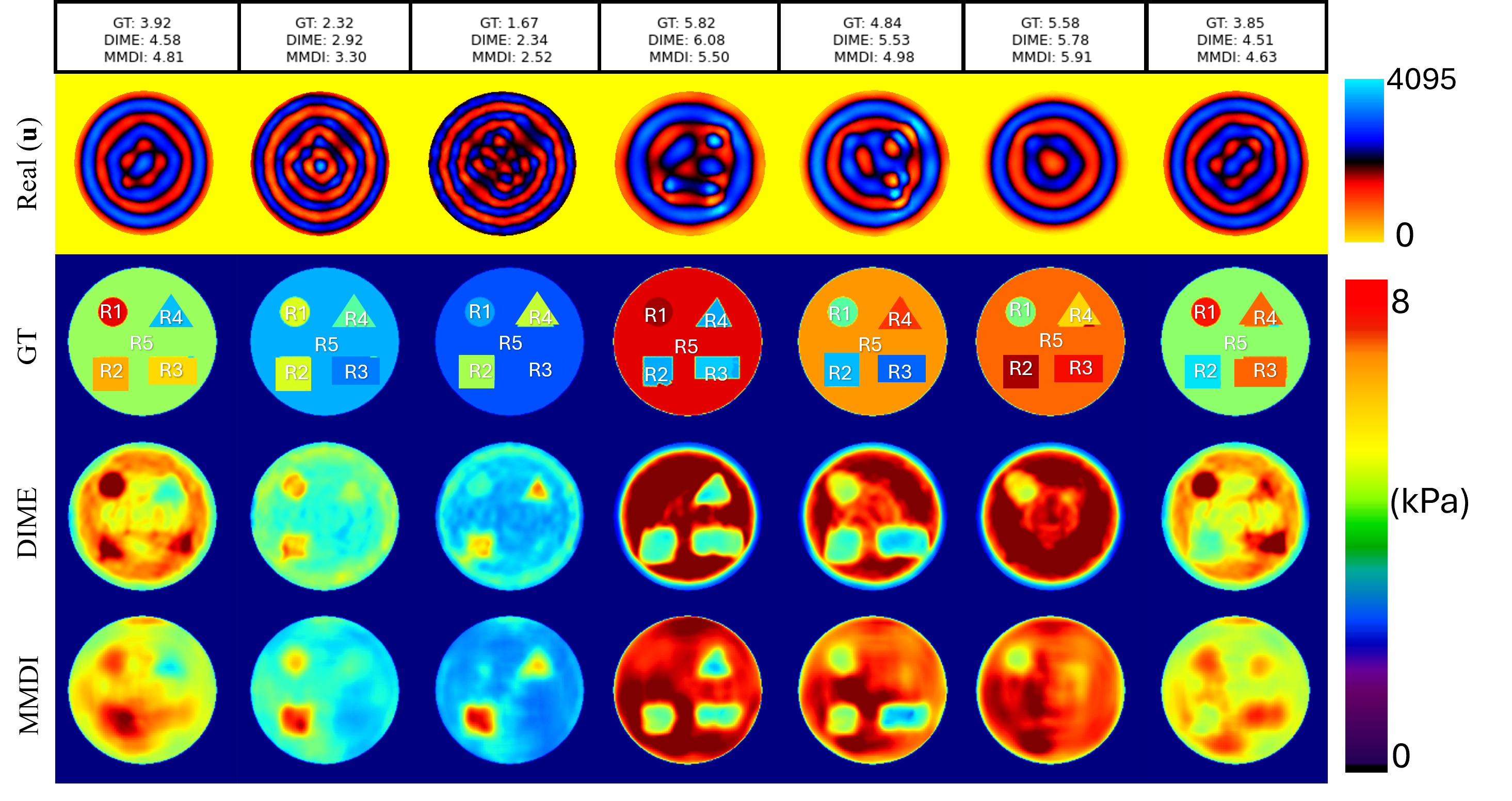}
    \caption{
        Evaluation of DIME on simulated heterogeneous phantoms.
        Representative examples of real wavefield components and the
        corresponding stiffness maps reconstructed using DIME and MMDI.
        Mean stiffness values (in kPa) are shown above each column.
    }
    \label{fig:yourlabel5}
\end{figure}

new page

\begin{figure}[htbp]
    \centering
    \includegraphics[width=\textwidth]{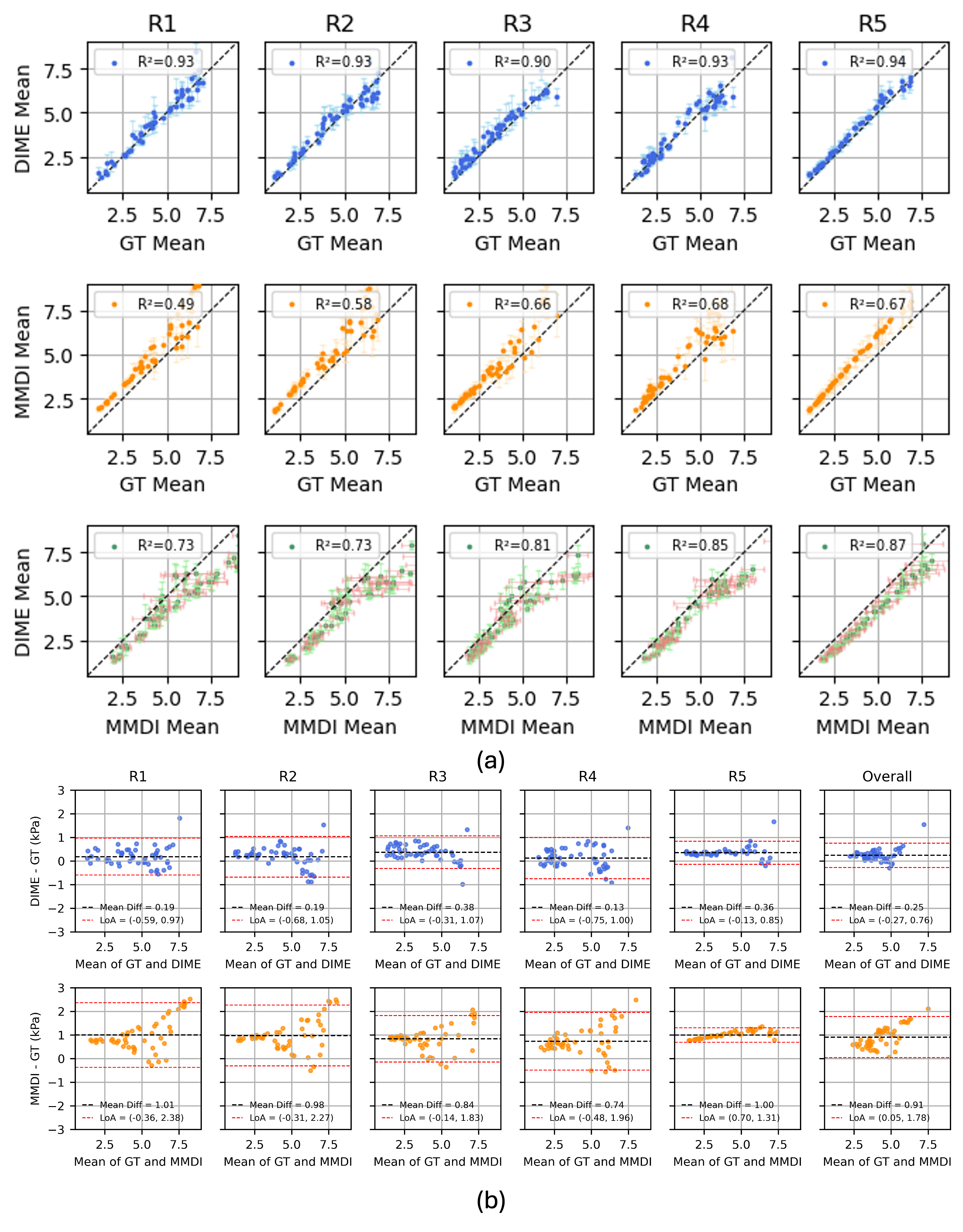}
    \caption{
        Region-wise correlation analysis across five test
phantoms (R1--R5). Top row: DIME vs. ground truth (GT); Middle row: MMDI
vs. GT; Bottom row: DIME vs. MMDI. DIME consistently shows stronger
agreement with GT across all regions (R² \textgreater{} 0.94), while
MMDI exhibits lower accuracy and higher variability. \textbf{(b)}
Bland--Altman plots comparing DIME (top) and MMDI (bottom) against
ground truth across five regions and overall. DIME shows lower bias and
tighter agreement limits than MMDI.
    }
    \label{fig:yourlabel6}
\end{figure}

\hypertarget{validation-of-dime-anatomy-informed-phantoms}{%
\subsection{4.2. Validation of DIME Anatomy Informed
Phantoms}\label{validation-of-dime-anatomy-informed-phantoms}}

To assess reconstruction accuracy, representative slices from the 
simulated liver dataset are shown in Fig.~7(a), including the real and 
imaginary displacement fields and the corresponding stiffness maps from 
DIME, MMDI, and GT. Visual inspection shows that DIME closely replicates 
the spatial stiffness patterns observed in GT, whereas MMDI tends to 
underestimate stiffness under the same reconstruction conditions. 
Quantitative analysis supports these findings. As shown in Fig.~7(b), 
DIME demonstrates a strong linear correlation with GT, with a Pearson’s 
correlation of $r = 0.988$, Spearman’s correlation of $\rho = 0.944$, and 
an ordinary least squares (OLS) coefficient of determination 
($R^2 = 0.977$), with data points closely aligned along the identity line 
($y = x$). In comparison, Fig.~7(c) shows that MMDI also correlates with 
GT ($r = 0.976$, $\rho = 0.889$), but with a lower regression slope 
($0.55$) and a smaller OLS $R^2 = 0.954$, indicating systematic 
underestimation. Bland--Altman analysis (Figs.~7(d--e)) further 
highlights these differences. DIME exhibits a small bias of $0.18 \,\text{kPa}$ 
and narrow limits of agreement ($-0.04 \,\text{kPa}$, $0.4 \,\text{kPa}$), 
indicating close agreement with GT. Conversely, MMDI shows a larger 
negative bias ($-0.46 \,\text{kPa}$) and wider limits of agreement 
($-0.94 \,\text{kPa}$, $0.03 \,\text{kPa}$), consistent with the observed 
underestimation.

\begin{figure}[htbp]
    \centering
    \includegraphics[width=\textwidth]{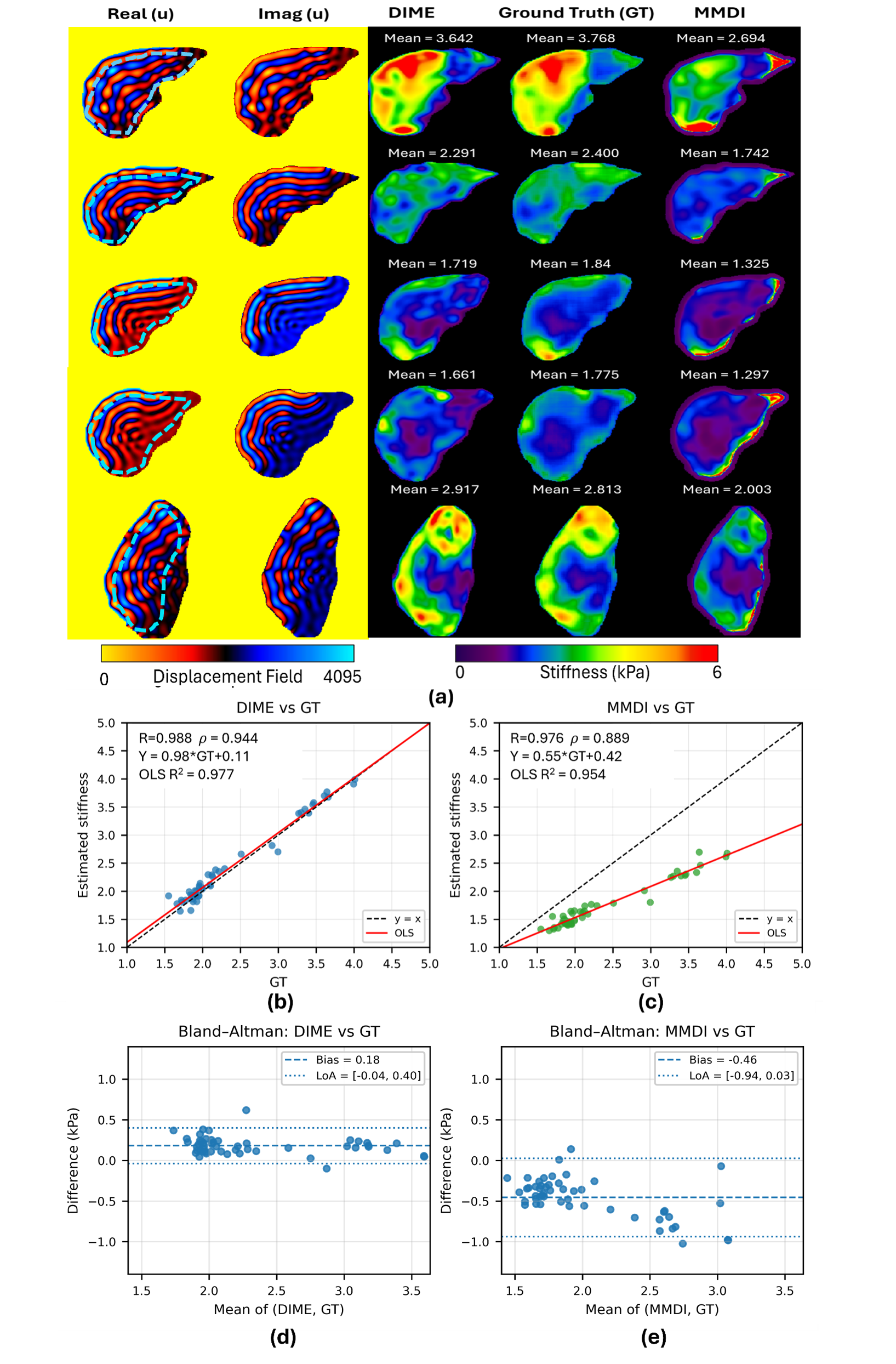}
    \caption{
        Evaluation of DIME on anatomy informed simulated
phantoms. \textbf{(a)} Representative examples showing the real and
imaginary components of the displacement field, together with stiffness
maps reconstructed using DIME, MMDI, and Ground Truth (GT). Mean
stiffness values are calculated within a region of interest (ROI)
defined by the 95\% confidence interval, with the ROI boundary indicated
on the real displacement field. \textbf{(b--c)} Statistical comparisons
of DIME and MMDI against GT, including multiple correlation
coefficients. \textbf{(d--e)} Bland--Altman analysis illustrating the
bias and limits of agreement for MMDI vs GT and DIME vs GT,
respectively.
    }
    \label{fig:yourlabel7}
\end{figure}

\subsection{4.3. Validation of DIME \emph{in vivo}}

Representative stiffness maps reconstructed from in vivo liver MRE data 
of five volunteers are shown in Fig.~8a. Visual inspection indicates that 
both DIME and MMDI capture comparable spatial patterns of stiffness within 
the liver. Quantitative comparison across 50 liver slices further supports 
this observation. As shown in Fig.~8b, DIME and MMDI demonstrated strong 
agreement, with Pearson's correlation coefficient of $r = 0.959$ and 
Spearman's correlation coefficient of $\rho = 0.866$. Linear regression 
analysis yielded a fit of $y = 0.97x + 0.69$, with a slope close to unity, 
indicating strong linear agreement between the two methods. However, the 
positive intercept suggests the presence of a bias, consistent with a 
tendency of MMDI to slightly underestimate stiffness relative to DIME.

\begin{figure}[htbp]
    \centering
    \includegraphics[width=\textwidth]{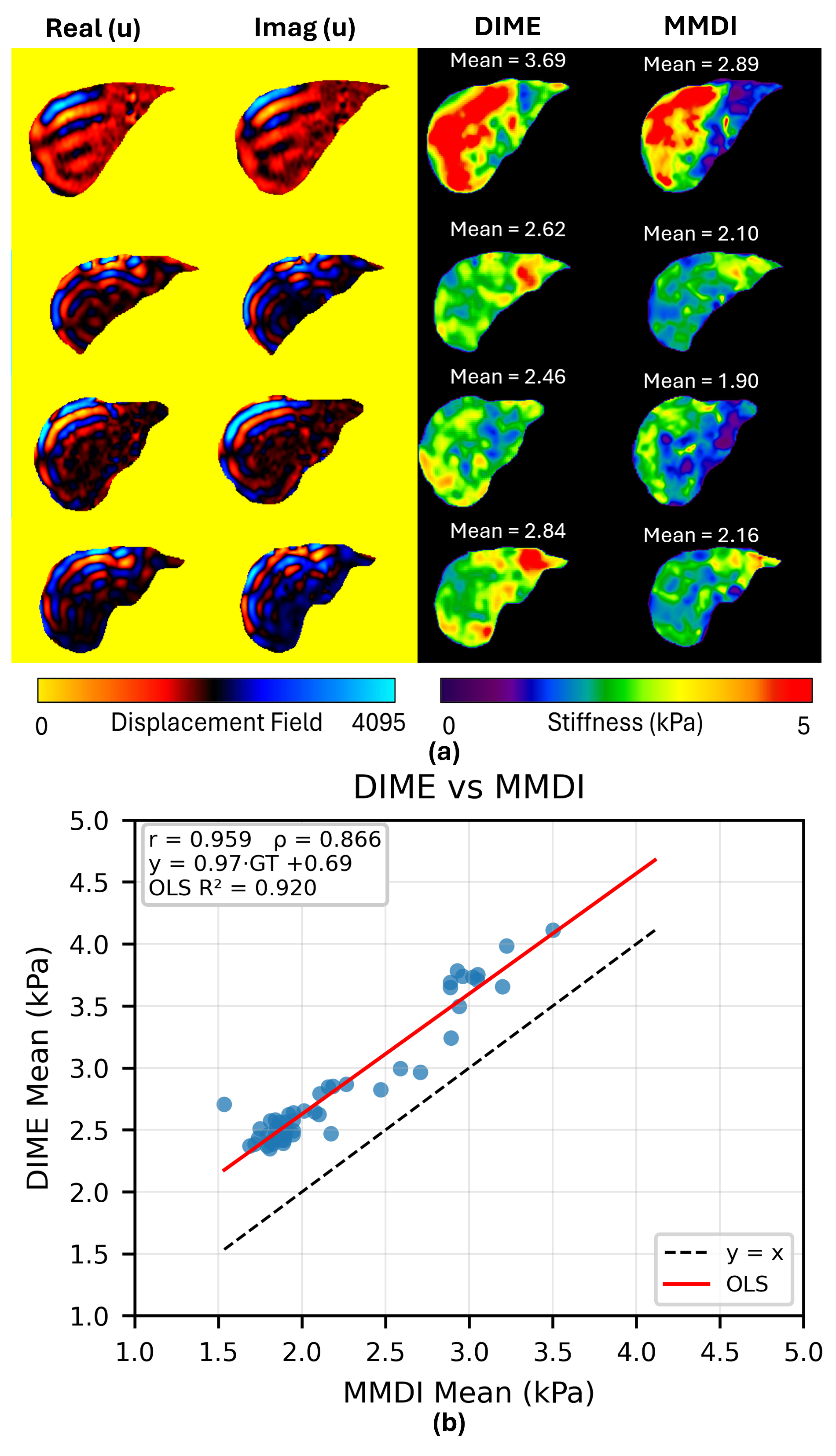}
    \caption{
        Reconstruction results of DIME and MMDI for in vivo liver data. 
        \textbf{(a)} Input in vivo wave images (left) with corresponding 
        stiffness maps for DIME and MMDI (right), annotated with mean ROI 
        stiffness values. 
        \textbf{(b)} Scatter plot comparing DIME and MMDI estimates, showing 
        strong agreement with Pearson's correlation coefficient 
        ($r = 0.959$) and Spearman's correlation coefficient 
        ($\rho = 0.866$). The regression line is given by 
        $y = 0.97x + 0.69$.
    }
    \label{fig:yourlabel8}
\end{figure}

\clearpage
\hypertarget{discussions}{%
\section{5. DISCUSSIONS}\label{discussions}}

In this work, we introduced DIME, a DL--based inversion framework that
leverages FEM-generated wavefields for stiffness reconstruction. Our
results across multiple validation studies demonstrated that DIME
recovers stiffness maps with higher fidelity to ground truth compared to
the commonly used MMDI. Importantly, the evaluation encompassed both
simple phantoms and anatomy informed FEM generated liver phantoms, as
well as in vivo liver MRE, allowing for a broad assessment of
performance.

\subsection{5.1. Comparison of DIME and MMDI on simple simulated phantoms}

In homogeneous phantom experiments, DIME maintained close agreement 
with ground truth, while MMDI consistently overestimated stiffness. 
Although both methods showed correlation with true values, MMDI 
displayed systematic bias. In heterogeneous phantoms, DIME more 
accurately delineated sharp stiffness boundaries and produced smoother 
reconstructions, consistent with the loss function’s penalization of 
inter-pixel variability. Region-based analysis revealed that DIME 
achieved higher $R^2$ values relative to ground truth across all ROIs, 
supported by Bland--Altman analysis showing narrower limits of agreement. 
By contrast, MMDI exhibited greater variability and a consistent bias 
toward overestimation. These findings suggest that DIME is particularly 
well-suited for detecting localized regions of elevated stiffness, such 
as those associated with tumors, where precise boundary recovery is 
clinically important.

\hypertarget{comparison-of-dime-and-mmdi-on-anatomy-informed-phantoms}{%
\subsection{\texorpdfstring{5.2. Comparison of DIME and MMDI on anatomy
informed phantoms
}{5.2. Comparison of DIME and MMDI on anatomy informed phantoms }}\label{comparison-of-dime-and-mmdi-on-anatomy-informed-phantoms}}

The anatomy-informed phantom study provided an intermediate step between
simple phantoms and in vivo data, preserving ground-truth accessibility
while introducing spatial heterogeneity. In this setting, DIME again
demonstrated higher agreement with the ground truth, reflecting higher
R² values and reduced bias compared to MMDI. While MMDI achieved
reasonable Pearson and Spearman correlations, the presence of bias
lowered its overall accuracy. An interesting trend emerged: while MMDI
overestimated stiffness in simple phantom experiments, it underestimated
stiffness in anatomy informed cases. This discrepancy likely arose from
the increased structural complexity of the waves in the anatomically
realistic stiffness distributions. These results demonstrate DIME's
ability to accurately recover global stiffness under idealized,
spatially heterogeneous conditions, outperforming the conventional MMDI.
To the best of our knowledge, this study represents one of the first
efforts where anatomically realistic heterogeneous stiffness maps were
used to benchmark DL--based inversion against conventional MMDI. This
setting provides a rare opportunity to assess reconstruction performance
under realistic wave propagation conditions while retaining access to
ground truth stiffness. Such conditions are seldom available in existing
datasets, making this comparison particularly valuable.

\hypertarget{comparison-of-dime-and-mmdi-on-in-vivo-liver-mre}{%
\subsection{\texorpdfstring{5.3. Comparison of DIME and MMDI on \emph{in
vivo} liver
MRE}{5.3. Comparison of DIME and MMDI on in vivo liver MRE}}\label{comparison-of-dime-and-mmdi-on-in-vivo-liver-mre}}

For \emph{in vivo} liver MRE data, the overall trend was consistent with
the phantom studies. Both DIME and MMDI produced stiffness maps that
captured the main spatial patterns and showed strong correlation.
However, MMDI tends to underestimate the stiffness values compared to
DIME, like the trend observed in FEM-generated anatomy informed phantom.

A similar bias pattern was also reported previous AI-based inversion
studies,\textsuperscript{27} where AI-based reconstructed stiffness maps
exhibited systematic bias relative to direct inversion results. However,
this bias was not analyzed in depth, likely due to the absence of
quantitative ground-truth stiffness values. This underscores the
importance of having access to realistic labeled datasets, as
demonstrated in our Study 3.2, which enabled a controlled evaluation and
direct comparison between the conventional MMDI and the proposed DIME
framework.

In our observation, this bias in MMDI inversion is likely linked to the
use of directional band-pass filtering as a preprocessing step. The
choice of cutoff frequencies in this filtering step (usually set between
2 and 128 waves per field of view in clinical liver MRE) has a strong
influence on the quantitative outcome. While these cutoff values are
commonly used in clinical practice, they may not necessarily be optimal
and can introduce bias by affecting the absolute stiffness values.
Although changing these parameters does not affect the overall spatial
appearance of the maps, it can shift the absolute stiffness values and
introduce bias. In contrast, DIME does not rely on such parameter
tuning, which makes its estimates more stable and less sensitive to
filtering choices.

\hypertarget{limitations}{%
\subsection{5.4. Limitations}\label{limitations}}

\emph{5.4.1. Noisy Measurements}

In this study, in vivo liver MRE data exhibited relatively low noise
levels. While noise was artificially introduced into the phantom
training data to encourage generalization, a systematic evaluation of
DIME and MMDI performance across varying noise levels was not conducted.
Future work will include controlled experiments at different noise
intensities to better characterize robustness.

\emph{5.4.2. Network Architecture}

A U-Net--based architecture was adopted as a baseline. However, other
network designs that have demonstrated success in medical imaging could
be explored. Evaluating advanced architectures, such as attention-based
networks or physics-driven loss functions,\textsuperscript{26} may
further improve inversion performance and generalizability.

\emph{5.4.3. Other directions of wave propagation}

The current model was trained on slice-by-slice phantom data using only
in-plane wave information. This restriction could lead to biased
stiffness estimation in complex geometries. Extending the framework to
fully 3D MRE acquisitions, where both in-plane and through-plane
displacement fields are available, would allow for more robust stiffness
reconstruction. Such an extension would also broaden the applicability
of DIME to challenging domains such as cardiovascular and
cardiopulmonary MRE.

\emph{5.4.4. Applications}

This work focused on liver MRE, an established clinical application for
staging fibrosis. While this provided a clinically relevant test case,
the DIME framework is not limited to hepatic applications. Future
studies will adapt the pipeline to other organs in different disease
states, incorporating 3D data, higher noise levels, and diverse wave
propagation patterns.

\emph{5.4.5. Study Population}

Validation was performed on a limited cohort of patients with fibrotic
liver conditions. Larger-scale clinical studies are needed to evaluate
DIME's reliability across different stages of fibrosis and to confirm
its potential for integration into routine clinical practice.

\hypertarget{conclusions}{%
\section{6. CONCLUSIONS}\label{conclusions}}

In this study, we introduced DIME, a DL--based inversion framework
trained on FEM-generated datasets to enable robust estimation of tissue
stiffness. The proposed pipeline provided a systematic comparison
against the conventional clinical MMDI approach across multiple
evaluation settings. In both simple and heterogeneous digital phantoms,
DIME demonstrated stronger correlations with ground truth. For anatomy
informed phantoms, DIME preserved fine spatial detail and achieved
higher correlations with ground truth relative to MMDI. Finally,
preliminary in vivo liver MRE experiments indicated that DIME could
generate reliable stiffness maps while mitigating some limitations
inherent to MMDI. Collectively, these findings highlight the potential
of DIME as a complementary inversion technique for clinical MRE,
warranting further validation in larger and more diverse patient
cohorts.

\hypertarget{acknowledgment}{%
\section{7. ACKNOWLEDGMENT}\label{acknowledgment}}

This work was supported by the National Institutes of Health (NIH) under
grant R01 AR075062 from the National Institute of Arthritis and
Musculoskeletal and Skin Diseases (NIAMS).

\clearpage
\section{8. REFERENCES}\label{references}

1. Duck, F. A. \emph{Physical Properties of Tissue: A Comprehensive Reference Book}. 
6th ed. Academic Press. 
Available at: \url{http://books.google.com/books?hl=en&lr=&id=UaXpAgAAQBAJ&oi=fnd&pg=PP1&ots=I6mT5dAlqi&sig=pm4iLHN_X0akcBYpcgRBI5f8MqU#v=onepage&q&f=false}

2. Duck FA. Mechanical Properties of Tissue. In: \emph{Physical
Properties of Tissues}. Elsevier; 1990:137-165.
doi:10.1016/B978-0-12-222800-1.50009-7

3. Venkatesh SK, Yin M, Ehman RL. Magnetic Resonance Elastography of
Liver: Clinical Applications. \emph{J Comput Assist Tomogr}.
2013;37(6):887-896. doi:10.1097/RCT.0000000000000032

4. Imajo K, Honda Y, Kobayashi T, et al. Direct Comparison of US and MR
Elastography for Staging Liver Fibrosis in Patients With Nonalcoholic
Fatty Liver Disease. \emph{Clin Gastroenterol Hepatol}.
2022;20(4):908-917.e11. doi:10.1016/j.cgh.2020.12.016

5. Elias D, Sideris L, Pocard M, et al. Incidence of Unsuspected and
Treatable Metastatic Disease Associated With Operable Colorectal Liver
Metastases Discovered Only at Laparotomy (and Not Treated When
Performing Percutaneous Radiofrequency Ablation). \emph{Ann Surg Oncol}.
2005;12(4):298-302. doi:10.1245/ASO.2005.03.020

6. McKnight AL, Kugel JL, Rossman PJ, Manduca A, Hartmann LC, Ehman RL.
MR Elastography of Breast Cancer: Preliminary Results. \emph{Am J
Roentgenol}. 2002;178(6):1411-1417. doi:10.2214/ajr.178.6.1781411

7. McKnight AL, Kugel JL, Rossman PJ, Manduca A, Hartmann LC, Ehman RL.
MR Elastography of Breast Cancer: Preliminary Results. \emph{Am J
Roentgenol}. 2002;178(6):1411-1417. doi:10.2214/ajr.178.6.1781411

8. Wang Y, Zhou J, Lin H, et al. Viscoelastic parameters derived from
multifrequency MR elastography for depicting hepatic fibrosis and
inflammation in chronic viral hepatitis. \emph{Insights Imaging}.
2024;15(1):91. doi:10.1186/s13244-024-01652-5

9. Bartsch KM, Schleip R, Zullo A, Hoppe K, Klingler W. The Stiffness
Comparison Test: A pilot study to determine inter-individual differences
in palpatory skill related to gender, age, and occupation-related
experience. \emph{J Bodyw Mov Ther}. 2020;24(4):1-6.
doi:10.1016/j.jbmt.2020.06.009

10. Muthupillai R, Lomas DJ, Rossman PJ, Greenleaf JF, Manduca A, Ehman
RL. Magnetic Resonance Elastography by Direct Visualization of
Propagating Acoustic Strain Waves. \emph{Science}.
1995;269(5232):1854-1857. doi:10.1126/science.7569924

11. Wymer DT, Patel KP, Burke WF, Bhatia VK. Phase-Contrast MRI:
Physics, Techniques, and Clinical Applications. \emph{RadioGraphics}.
2020;40(1):122-140. doi:10.1148/rg.2020190039

12. Moran PR. A flow velocity zeugmatographic interlace for NMR imaging
in humans. \emph{Magn Reson Imaging}. 1982;1(4):197-203.
doi:10.1016/0730-725X(82)90170-9

13. Manduca A, Oliphant TE, Dresner MA, et al. Magnetic resonance
elastography: Non-invasive mapping of tissue elasticity. \emph{Med Image
Anal}. 2001;5(4):237-254. doi:10.1016/S1361-8415(00)00039-6

14. Papazoglou S, Hamhaber U, Braun J, Sack I. Algebraic Helmholtz
inversion in planar magnetic resonance elastography. \emph{Phys Med
Biol}. 2008;53(12):3147-3158. doi:10.1088/0031-9155/53/12/005

15. Romano AJ, Bucaro JA, Ehnan RL, Shirron JJ. Evaluation of a material
parameter extraction algorithm using MRI-based displacement
measurements. \emph{IEEE Trans Ultrason Ferroelectr Freq Control}.
2000;47(6):1575-1581. doi:10.1109/58.883546

16. Manduca A, Muthupillai R, Rossman PJ, Greenleaf JF, Ehman RL. Local
wavelength estimation for magnetic resonance elastography. In:
\emph{Proceedings of 3rd IEEE International Conference on Image
Processing}. Vol 3. IEEE; 1996:527-530. doi:10.1109/ICIP.1996.560548

17. Knutsson H, Westin CF, Granlund G. Local multiscale frequency and
bandwidth estimation. In: \emph{Proceedings of 1st International
Conference on Image Processing}. Vol 1. IEEE Comput. Soc. Press;
1994:36-40. doi:10.1109/ICIP.1994.413270

18. Silva AM, Grimm RC, Glaser KJ, et al. Magnetic resonance
elastography: evaluation of new inversion algorithm and quantitative
analysis method. \emph{Abdom Imaging}. 2015;40(4):810-817.
doi:10.1007/s00261-015-0372-5

19. Van Houten EEW, Paulsen KD, Miga MI, Kennedy FE, Weaver JB. An
overlapping subzone technique for MR-based elastic property
reconstruction. \emph{Magn Reson Med}. 1999;42(4):779-786.
doi:10.1002/(SICI)1522-2594(199910)42:4\%3C779::AID-MRM21\%3E3.0.CO;2-Z

20. Reeder SB. Emergence of 3D MR Elastography--based Quantitative
Markers for Diffuse Liver Disease. \emph{Radiology}.
2021;301(1):163-165. doi:10.1148/radiol.2021211444

21. Fu Y, Lei Y, Wang T, Curran WJ, Liu T, Yang X. Deep learning in
medical image registration: a review. \emph{Phys Med Biol}.
2020;65(20):20TR01. doi:10.1088/1361-6560/ab843e

22. Rayed MdE, Islam SMS, Niha SI, Jim JR, Kabir MM, Mridha MF. Deep
learning for medical image segmentation: State-of-the-art advancements
and challenges. \emph{Inform Med Unlocked}. 2024;47:101504.
doi:10.1016/j.imu.2024.101504

23. Cai L, Gao J, Zhao D. A review of the application of deep learning
in medical image classification and segmentation. \emph{Ann Transl Med}.
2020;8(11):713-713. doi:10.21037/atm.2020.02.44

24. Heckel R, Jacob M, Chaudhari A, Perlman O, Shimron E. Deep Learning
for Accelerated and Robust MRI Reconstruction: a Review. \emph{arXiv}.
Preprint posted online April 24, 2024. doi:10.48550/arXiv.2404.15692

25. Murphy MC, Manduca A, Trzasko JD, Glaser KJ, Huston J, Ehman RL.
Artificial neural networks for stiffness estimation in magnetic
resonance elastography: Neural Network Inversion for MRE. \emph{Magn
Reson Med}. 2018;80(1):351-360. doi:10.1002/mrm.27019

26. Solamen L, Shi Y, Amoh J. Dual Objective Approach Using A
Convolutional Neural Network for Magnetic Resonance Elastography.
\emph{arXiv}. Preprint posted online 2018. doi:10.48550/ARXIV.1812.00441

27. Scott JM, Arani A, Manduca A, et al. Artificial neural networks for
magnetic resonance elastography stiffness estimation in inhomogeneous
materials. \emph{Med Image Anal}. 2020;63:101710.
doi:10.1016/j.media.2020.101710

28. Ma S, Wang R, Qiu S, et al. MR Elastography With Optimization-Based
Phase Unwrapping and Traveling Wave Expansion-Based Neural Network
(TWENN). \emph{IEEE Trans Med Imaging}. 2023;42(9):2631-2642.
doi:10.1109/TMI.2023.3261346

29. Ragoza M, Batmanghelich K. Physics-Informed Neural Networks for
Tissue Elasticity Reconstruction in Magnetic Resonance Elastography. In:
Greenspan H, Madabhushi A, Mousavi P, et al., eds. \emph{Medical Image
Computing and Computer Assisted Intervention -- MICCAI 2023}. Vol 14229.
Lecture Notes in Computer Science. Springer Nature Switzerland;
2023:333-343. doi:10.1007/978-3-031-43999-5\_32

30. Zhang J, Mu X, Lin X, et al. Quantification of tissue stiffness with
magnetic resonance elastography and finite difference time domain (FDTD)
simulation-based spatiotemporal neural network. \emph{Magn Reson
Imaging}. 2025;118:110353. doi:10.1016/j.mri.2025.110353

31. Bustin H, Meyer T, Reiter R, et al. ElastoNet: Neural network-based
multicomponent MR elastography wave inversion with uncertainty
quantification. \emph{Med Image Anal}. 2025;105:103642.
doi:10.1016/j.media.2025.103642

32. Ulyanov D, Vedaldi A, Lempitsky V. Deep Image Prior. \emph{Int J
Comput Vis}. 2020;128(7):1867-1888. doi:10.1007/s11263-020-01303-4

33. Iftikhar H, Ahmad R, Kolipaka A. Curating dataset for AI -based
Stiffness Estimation in MR Elastography Using Finite Element Modeling
and Polynomial Curve Fitting.
https://archive.ismrm.org/2025/4000\_8x8Z7BRJ4.html

34. Kim YS, Jang YN, Song JS. Comparison of gradient-recalled echo and
spin-echo echo-planar imaging MR elastography in staging liver fibrosis:
a meta-analysis. \emph{Eur Radiol}. 2018;28(4):1709-1718.
doi:10.1007/s00330-017-5149-5

35. Wagner M, Besa C, Bou Ayache J, et al. Magnetic Resonance
Elastography of the Liver: Qualitative and Quantitative Comparison of
Gradient Echo and Spin Echo Echoplanar Imaging Sequences. \emph{Invest
Radiol}. 2016;51(9):575-581. doi:10.1097/RLI.0000000000000269

\end{document}